\documentclass[11pt,a4paper]{article}
\usepackage[hyperref]{emnlp-ijcnlp-2019}
\usepackage{times}
\usepackage{latexsym}
\usepackage{microtype}
\usepackage{booktabs}
\usepackage{url}
\usepackage{eurosym}
\usepackage{todonotes}
\usepackage{multirow}
\usepackage{subcaption}
\usepackage{amssymb}
\usepackage{amsmath}
\usepackage{array}
\usepackage{soul}
\usepackage{xcolor}
\usepackage{enumitem}

\usepackage{url}

\aclfinalcopy 

\newif\ifcomments
\commentstrue 
\ifcomments
    \newcommand{\dheeru}[1]{{\color{orange}({\bf DD}: #1)}}
    \newcommand{\alon}[1]{{\color{olive}({\bf AT}: #1)}}
    \newcommand{\sameer}[1]{{\color{red}({\bf SS}: #1)}}
    \newcommand{\matt}[1]{{\color{blue}({\bf MG}: #1)}}
    \newcommand{\ananth}[1]{{\color{blue}({\bf AG}: #1)}}
\else
    \providecommand{\dheeru}[1]{}
    \providecommand{\alon}[1]{}
    \providecommand{\sameer}[1]{}
    \providecommand{\matt}[1]{}
    \providecommand{\ananth}[1]{}
\fi

\title{ORB: An Open Reading Benchmark \\ for Comprehensive Evaluation of Machine Reading Comprehension }
\author{Dheeru Dua\textsuperscript{$\clubsuit$}, Ananth Gottumukkala\textsuperscript{$\clubsuit$}, Alon Talmor\textsuperscript{$\heartsuit$}, \\ \textbf{Sameer Singh}\textsuperscript{$\clubsuit$}, and \textbf{Matt Gardner}\textsuperscript{$\spadesuit$} \\
  \textsuperscript{$\clubsuit$}University of California, Irvine, USA \\
  \textsuperscript{$\heartsuit$}Allen Institute for Artificial Intelligence, Seattle, Washington, USA \\
  \textsuperscript{$\spadesuit$}Allen Institute for Artificial Intelligence, Irvine, California, USA \\
  {\tt ddua@uci.edu} }

\date{}

\begin{document}
\maketitle

\begin{abstract}
Reading comprehension is one of the crucial tasks for furthering research in natural language understanding. A lot of diverse reading comprehension datasets have recently been introduced to study various phenomena in natural language, ranging from simple paraphrase matching and entity typing to entity tracking and understanding the implications of the context. Given the availability of many such datasets, comprehensive and reliable evaluation is tedious and time-consuming for researchers working on this problem. We present an evaluation server, \textbf{ORB}, that reports performance on seven diverse reading comprehension datasets, encouraging and facilitating testing a single model's capability in understanding a wide variety of reading phenomena. The evaluation server places no restrictions on how models are trained, so it is a suitable test bed for exploring training paradigms and representation learning for general reading facility. As more suitable datasets are released, they will be added to the evaluation server.  We also collect and include synthetic augmentations for these datasets, testing how well models can handle out-of-domain questions.




\end{abstract}

\section{Introduction}
Research in reading comprehension, the task of answering questions about a given passage of text, has seen a huge surge of interest in recent years, with many large datasets introduced targeting various aspects of reading \cite{rajpurkar2016squad,dua2019drop,dasigi2019quoref,Lin2019ReasoningOP}.  However, as the number of datasets increases, evaluation on all of them becomes challenging, encouraging researchers to overfit to the biases of a single dataset.  Recent research, including MultiQA \citep{Talmor2019MultiQAAE} and the MRQA workshop shared task, aim to facilitate training and evaluating on several reading comprehension datasets at the same time.  To further aid in this direction, we present an evaluation server that can test a single model across many different datasets, including on their hidden test sets in some cases.  We focus on datasets where the core problem is natural language understanding, not information retrieval; models are given a single passage of text and a single question and are required to produce an answer.

As our goal is to provide a broad suite of questions that test a single model's reading ability, we additionally provide synthetic augmentations to some of the datasets in our evaluation server.  Several recent papers have proposed question transformations that result in out-of-distribution test examples, helping to judge the generalization capability of reading models \cite{ribeiro2018semantically,ribeiro2019red,zhu-etal-2019-learning}.  We collect the best of these, add some of our own, and keep those that generate reasonable and challenging questions.  We believe this strategy of evaluating on many datasets, including distribution-shifted synthetic examples, will lead the field towards more robust and comprehensive reading comprehension models.

Code for the evaluation server, including a script to run it on the dev sets of these datasets and a leaderboard showing results on their hidden tests, can be found at \url{https://leaderboard.allenai.org/orb}

\section{Datasets}

We selected seven existing datasets that target various complex linguistic phenomena such as coreference resolution, entity and event detection, etc., capabilities which are desirable when testing a model for reading comprehension.
We chose datasets that adhere to two main properties: First, we exclude from consideration any multiple choice dataset, as these typically require very different model architectures, and they often have biases in how the distractor choices are generated. Second, we require that the dataset be originally designed for answering isolated questions over a single, given passage of text.  We are focused on evaluating \emph{reading} performance, not \emph{retrieval}; reading a single passage of text is far from solved, so we do not complicate things by adding in retrieval, conversation state, or other similar complexities.

It is our intent to add to the evaluation server any high-quality reading comprehension dataset that is released in the future that matches these restrictions.

We now briefly describe the datasets that we include in the initial release of ORB.  Table \ref{tab:dataset} gives summary statistics of these datasets. Except where noted, we include both the development and test sets (including hidden test sets) in our evaluation server for all datasets.

\paragraph{SQuAD \textnormal{\cite{rajpurkar2016squad}}} requires a model to perform lexical matching between the context and the question to predict the answer. This dataset provides avenues to learn predicate-argument structure and multi-sentence reasoning to some extent. It was collected by asking crowd-workers to create question-answer pairs from Wikipedia articles such that the answer is a single-span in the context. The dataset was later updated to include unanswerable questions \cite{rajpurkar2018squadrun}, giving a harder question set without as many reasoning shortcuts. We include only the development sets of SQuAD 1.1 and SQuAD 2.0 in our evaluation server.

\paragraph{DuoRC \textnormal{\cite{Sankaranarayanan2019duorc}}} tests if the model can generalize to answering semantically similar but syntactically different paraphrased questions. The questions are created on movie summaries obtained from two sources, Wikipedia and IMDB. The crowd-workers formalized questions based on Wikipedia contexts and in turn answered them based on the IMDB context. This ensured that the model will not rely solely on lexical matching, but rather utilize semantic understanding. The answer can be either a single-span from context or free form text written by the annotator. 

\paragraph{Quoref \textnormal{\cite{dasigi2019quoref}}} focuses on understanding coreference resolution, a challenging aspect of natural language understanding. It helps gauge how a model can handle ambiguous entity and event resolution to answer a question correctly. This dataset was created by asking crowd workers to write questions and multi-span answers from Wikipedia articles that centered around pronouns in the context.

\paragraph{DROP \textnormal{\cite{dua2019drop}}} attempts to force models to have a more comprehensive understanding of a paragraph, by constructing questions that query many parts of the paragraph at the same time. These questions involve reasoning operations that are mainly rudimentary mathematical skills such as addition, subtraction, maximum, minimum, etc. To perform well on this dataset a model needs to locate multiple spans associated with questions in the context and perform a set of operations in a hierarchical or sequential manner to obtain the answer. The answer can be either a set of spans from the context, a number or a date.

\paragraph{ROPES \textnormal{\cite{Lin2019ReasoningOP}}} centers around understanding the implications of a passage of text, particularly dealing with the language of causes and effects.  A system is given a background passage, perhaps describing the effects of deforestation on local climate and ecosystems, and a grounded situation involving the knowledge in the background passage, such as, \emph{City A has more trees than City B}.  The questions then require grounding the effects described in the background, perhaps querying which city would more likely have greater ecological diversity.  This dataset can be helpful in understanding how to apply the knowledge contained in a passage of text to a new situation. 

\paragraph{NewsQA \textnormal{\cite{trischler2017newsqa}}} dataset focuses on paraphrased questions with predicate-argument structure understanding. To some extent it is similar to DuoRC, however the examples are collected from news articles and offers diverse linguistic structures. This crowd-sourced dataset was created by asking annotators to write questions from CNN/DailyMail articles as context.

\paragraph{NarrativeQA \textnormal{\cite{kocisky2018narrativeqa}}} focuses on understanding temporal reasoning among various events that happen in a given movie plot. It also tests the model’s ability to ``hop'' between various parts of the context and not rely solely on sequential reasoning. The dataset was constructed by aligning books from Gutenberg~\footnote{\url{http://www.gutenberg.org/}} with the summaries of their movie adaptations from various web resources. The crowd workers were asked to create complex questions about characters, narratives, events etc. from summaries and typically can be answered from summaries. In addition, crowd workers were required to provide answers that do not have high overlap with the context. In accordance with our format, we only use the version with the summaries as context in our evaluation server.

\begin{table}[]
    \centering
    \footnotesize
    \begin{tabular}{p{1.5cm}p{0.75cm}p{0.75cm}p{0.75cm}p{0.75cm}}
    \toprule
     \textbf{Dataset} &  \textbf{Dev Size} &  \textbf{Test Size} &  \textbf{Context Length (Avg)} &  \textbf{Answer Length (Avg)}\\
     \midrule
     SQuAD1.1 & 10,570 & - & 123.7 & 4.0\\
     SQuAD2.0 & 10,570 & - & 127.5 & 4.2 \\
     DuoRC & 12,233 & 13,449 & 1113.6 & 2.8 \\
     Quoref & 2,418 & 2,537 & 348.2 & 2.7 \\
     DROP & 9,536 & 9,622 & 195.1 & 1.5 \\
     ROPES & 1,204 & 1,015 & 177.1 & 1.2\\
     NewsQA & 5,166 & 5,126 & 711.3  & 5.1\\
     NarrativeQA & 3,443 & 10,557 & 567.9 & 4.7\\
     \bottomrule
    \end{tabular}
    \caption{Dataset Statistics}
    \label{tab:dataset}
\end{table}

\section{Synthetic Augmentations}



Prior works~\cite{jia2017adversarial} have shown that RC models are brittle to minor perturbations in original dataset. Hence, to test the model's ability to generalize to out-of-domain syntactic structures and be logically consistent in its answers, we automatically generate questions based on various heuristics. These heuristics fall in two broad categories. 
\begin{enumerate}
    \item The question is paraphrased to a minimal extent to create new syntactic structures, keeping the semantics of the question largely intact and without making any changes to the original context and answer.
    \item The predicate-argument structures of a given question-answer pair are leveraged to create new WH-questions based on the object in the question instead of the subject. This rule-based method, adopted from \cite{ribeiro2019red}, changes the question and answer keeping the context fixed.
\end{enumerate}



We use five augmentation techniques, where the first four techniques fall into the first category and the last technique falls into the second category.

\paragraph{Invert Choice} transforms a binary choice question by changing the order in which the choices are presented, keeping the answer the same. 
\paragraph{More Wrong Choice} transforms a binary choice question by substituting the wrong choice in the question with another wrong choice from the passage.
\paragraph{No Answer} substitutes a name in the question for a different name from the passage to create with high probability a new question with no answer. 
\paragraph{SEARs} creates minimal changes in word selection or grammar while maintaining the original meaning of the question according to the rules described by \citet{ribeiro2018semantically}.
\paragraph{Implication} creates a new question-answer pair, where the object of the original question is replaced with the answer directly resulting in creation of a new WH-question where the answer is now the object of the original question. These transformations are performed based on rules described by \citet{ribeiro2019red}. 
\medskip

We attempted all the above augmentation techniques on all the datasets (except NarrativeQA where entity and event tracking is complex and these simple transformations can lead to a high number of false positives). Table~\ref{tab:augyield} shows the number of augmentations generated by each augmentation technique-dataset pair. A few sample augmentations are shown in Table~\ref{tab:main_examples}.

\begin{table}
    \small
    \centering
    \begin{tabular}{lrrrrr}
    \toprule
     \textbf{Dataset} & \textbf{IC} & \textbf{MWC} & \textbf{Imp} & \textbf{No-Ans} & \textbf{SEARs} \\
    \midrule
    NewsQA  & 0 & 0 & 501 & 347 & 16009  \\
    QuoRef  & 0 & 0 & 79 & 385 & 11759  \\
    DROP    & 1377 & 457 & 113 & 284 & 16382  \\
    SQuAD   & 16 & 0 & 875 & 594 & 28188  \\
    ROPES   & 637 & 119 & 0 & 201 & 2909  \\
    DuoRC   & 22 & 0 & 2706 & - & 45020  \\
    \bottomrule
    \end{tabular}
    \caption{Yields of augmented datasets}
    \label{tab:augyield}
\end{table}

\begin{table*}[!t]
\centering
\footnotesize
\resizebox{1.0\textwidth}{!}{
\begin{tabular}{p{2cm}p{7cm}p{3cm}p{3cm}}
\toprule
{\bf Template Type} & {\bf Context (truncated)} &  {\bf Original QA Pair} & {\bf Generated QA Pair} \\
 \midrule
 \textbf{Invert Choice} &  ... before halftime thanks to a David Akers 32-yard field goal, giving Detroit a 17-14 edge ... in the third, Washington was able to equalize with John Potter making his first career field goal from 43 yards out ... in the fourth, Detroit took the lead again, this time by way of Akers hitting a 28-yard field goal, giving Detroit a 20-17 lead... & \textbf{Q:} Which player scored more field goals, David Akers or John Potter?  \textbf{A:} David Akers & \textbf{Q:} Which player scored more field goals, John Potter or David Akers? \textbf{A:} David Akers \\
 \midrule
 \textbf{More Wrong Choice} & The first issue in 1942 consisted of denominations of 1, 5, 10 and 50 centavos and 1, 5, and 10 Pesos. ... 1944 ushered in a 100 Peso note and soon after an inflationary 500 Pesos note. In 1945, the Japanese issued a 1,000 Pesos note...  &  \textbf{Q:} Which year ushered in the largest Pesos note, 1944 or 1945? \hspace{1cm} \textbf{A:} 1945 &  \textbf{Q:} Which year ushered in the largest Pesos note, 1942 or 1945? \hspace{1cm} \textbf{A:} 1945 \\
 \midrule
 \textbf{Implication} & ... In 1562, naval officer Jean Ribault led an expedition that explored Florida and the present-day Southeastern U.S., and founded the outpost of Charlesfort on Parris Island, South Carolina... & \textbf{Q:} When did Ribault first establish a settlement in South Carolina? \hspace{2cm} \textbf{A:} 1562 & \textbf{Q:} Who established a settlement in South Carolina in 1562? \hspace{2cm} \textbf{A:} Ribault \\
 \midrule
 \textbf{No Answer} & From 1975, Flavin installed permanent works in Europe and the United States, including ... the Union Bank of Switzerland, Bern (1996). ... The 1930s church was designed by Giovanni Muzio... & \textbf{Q:} Which permanent works did Flavin install in 1996? \hspace{2cm} \textbf{A:} Union Bank of Switzerland, Bern & \textbf{Q:} Which permanent works did Giovanni Muzio install in 1996? \hspace{2cm} \textbf{A:} No Answer \\
 \midrule
 \textbf{SEARs} & ... Dhul-Nun al-Misri and Ibn Wahshiyya were the first historians to study hieroglyphs, by comparing them to the contemporary Coptic language used by Coptic priests in their time... & \textbf{Q:} What did historians compare to the Coptic language? \hspace{2cm} \textbf{A:} hieroglyphs & \textbf{Q:} What'd historians compare to the Coptic language? \hspace{2cm} \textbf{A:} hieroglyphs \\
 \bottomrule
\end{tabular}}
\caption{Examples of generated augmentations with various templates.}
\label{tab:main_examples}
\end{table*}


After generating all the augmented datasets, we manually identified the augmentation technique-dataset pairs which led to high-quality augmentations. We sample 50 questions from each augmented dataset and record whether they satisfy the three criteria given below.

\begin{enumerate}[nosep]
    \item Is the question understandable, with little to no grammatical errors?
    \item Is the question semantically correct?
    \item Is the new answer the correct answer for the new question?
\end{enumerate}


\begin{table}
    \small
    \centering
    \begin{tabular}{lrrrrr}
    \toprule
     \textbf{Dataset} & \textbf{IC} & \textbf{MWC} & \textbf{Imp} & \textbf{No-Ans} & \textbf{SEARs} \\
    \midrule
    NewsQA  & - & - & 47 & 47 & 50  \\
    QuoRef  & - & - & 45 & 48 & 50  \\
    DROP    & 46 & 42 & 36 & 48 & 50  \\
    SQuAD   & 15/16 & - & 47 & 48 & 50  \\
    ROPES   & 48 & 36 & - & 11 & 50  \\
    DuoRC   & 18/22 & - & 47 & - & 50  \\
    \bottomrule
    \end{tabular}
    \caption{Quality of augmented datasets (\# of good questions out of 50 sampled)}
    \label{tab:augquality}
\end{table}

Table~\ref{tab:augquality} shows the number of high-quality questions generated for each dataset. We keep the augmentation technique-dataset pairs where at least 90\% of the question-answer pairs satisfy the above three criteria. We further test the performance of these augmentations (Section 4) on a BERT~\cite{devlin2018bert} based model to establish if the dataset has a sufficiently different question distribution from the original and has enough independent value to be incorporated into the evaluation server.




\section{Experiments}
\subsection{Model}
We train a numerically-aware BERT-based model\footnote{https://github.com/raylin1000/drop-bert} (NABERT) on all the seven datasets and test its performance on existing datasets and synthetic augmentations. NABERT is a BERT based model with the ability to perform discrete operations like counting, addition, subtraction etc. We added support for ``impossible'' answers in the existing NABERT architecture by extending the answer type predictor which classifies the type of reasoning involved given a question into one of the following five categories:  {\emph{number}, \emph{span}, \emph{date}, \emph{count}, \emph{impossible}}. All the hyper-parameter settings were kept the same.

We noticed \emph{catastrophic forgetting} on randomly sampling a minibatch for training, from all the datasets pooled together. To alleviate this problem, we sampled uniformly from each dataset in the beginning and then switched to sampling in proportion to the size of each dataset towards the end of the epoch~\cite{pmlr-v97-stickland19a}. This helped improve the performance on several dataset by 3-4\% in EM, however, there is still a lot of room for improvement on this front. We also tried a simple BERT model~\cite{devlin2019bert} and MultiQA~\cite{Talmor2019MultiQAAE} but NABERT gave the best results on the seven development sets.

In case of DuoRC and NarrativeQA, some answers are free-form human generated and do not have an exact overlap with the context. However, the NABERT model is trained to predict a span's start and end indices in the context. So for answers, which are not exact spans from the context we pick a span which has the highest ROUGE-L with the gold answer to serve as labels for training. However, for evaluation we use the original gold answer and not the extracted passage span for evaluating the model's performance.

\subsection{Existing Dataset Performance}
Table~\ref{tab:performance} shows the result of evaluating on all of the development and test sets using our evaluation server. We chose the official metrics adopted by the individual datasets to evaluate the performance of our baseline model. As can be seen in the table, the results are quite poor, significantly below single-dataset state-of-the-art on all datasets.  The training of our initial baseline appears to be dominated by SQuAD 1.1, or perhaps SQuAD 1.1 mainly tests reasoning that is common to all of the other datasets.  Significant research is required to build reading systems and develop training regimes that are general enough to handle multiple reading comprehension datasets at the same time, even when all of the datasets are seen at training time.


\begin{table}
    \small
    \begin{minipage}{0.48\textwidth}
    \centering
    \begin{tabular}{lrrrr}
    \toprule
     \multirow{2}{*}{\bf Dataset}    & \multicolumn{2}{c}{\bf Dev}  & \multicolumn{2}{c}{\bf Test} \\
     \cmidrule(lr){2-3}
      \cmidrule(lr){4-5}
         & EM & F$_1$ & EM & F$_1$\\
    \midrule
    NewsQA   &   29.34 & 45.40  & 29.69 & 46.19 \\
    Quoref   &    34.49  &    42.65  & 30.13 & 38.39   \\
    DROP    &   19.09  &    23.16   & 17.69 & 21.87   \\
    SQuAD 1.1   &    68.03  &  78.55   & - & - \\
    SQuAD 2.0   &   33.70  &  39.17  & - & - \\
    ROPES   &    40.03  &   49.07   & 47.96 & 56.06  \\
    DuoRC   &  25.65  & 34.28  & 23.44  &   31.73   \\
    \bottomrule
    \end{tabular}
    \end{minipage}\vfill \vfill 
    \begin{minipage}{0.48\textwidth}
    \centering
    \begin{tabular}{p{1cm}p{0.75cm}p{0.75cm}p{1cm}p{1cm}}  \\
    \textbf{Narrative QA} & BLEU-1 & BLEU-4 & METEOR & ROUGE-L (F1) \\
    \midrule
      Dev Set &   0.17 & 0.021 & 0.33 & 0.52  \\
      Test Set &  0.16  & 0.019 & 0.33 & 0.53  \\
    \bottomrule
    \end{tabular}
    \end{minipage}
    \caption{Performance on baseline BERT model on different datasets  }
    \label{tab:performance}
\end{table}

\subsection{Synthetic Augmentations}
 Table~\ref{tab:error_analysis} shows the performance of the baseline model on various development sets and heuristically generated questions. The \textbf{More Wrong Choice} augmentation is omitted since a high enough quality and/or yield of questions could not be ensured for any of the datasets. When evaluated on out-of-domain linguistic structures, performance drops significantly for some augmentation-dataset pairs but only marginally for others. For questions generated by the \textbf{Invert Choice} augmentation, the model struggles to grasp the correct reasoning behind two answer options like \emph{Art Euphoric or Trescott Street} and changes the prediction when the choices are flipped. However, relative to the dev set performances on the original datasets, the performance drop is almost nonexistent. For the \textbf{SEARs} based augmentation the generated linguistic variations are close to in-domain syntactic structure so we do not see much performance drop in most of the datasets except for ROPES and NewsQA. The \textbf{Implication} style questions create a large performance drop for NewsQA and SQuAD while having a performance boost for DuoRC. Finally, the \textbf{No-Ans} type questions have the worst performance across board for all datasets.


\begin{table*}[!t]
\centering
\footnotesize
\begin{minipage}{\textwidth}
\centering
    \begin{tabular}{lrrrrrrrrrr}
    \toprule
     \multirow{2}{*}{\bf } & \multicolumn{2}{c}{\bf Dev} & 
     \multicolumn{2}{c}{\bf IC} & \multicolumn{2}{c}{\bf Imp} & \multicolumn{2}{c}{\bf No-Ans} & \multicolumn{2}{c}{\bf SEARs}  \\
     \cmidrule(lr){2-3}
     \cmidrule(lr){4-5}
     \cmidrule(lr){6-7}
     \cmidrule(lr){8-9}
     \cmidrule(lr){10-11}
         & EM & F$_1$ & EM & F$_1$ & EM & F$_1$ & EM & F$_1$ & EM & F$_1$ \\
    \midrule
    NewsQA  & 29.34 & 45.40 & - & - & 23.35 & 34.36 & 0.02 & 0.02 & 21.34 & 33.33  \\
    QuoRef  & 34.49 & 42.65 & - & -  & 32.91 & 44.84 & 0.0 & 0.0 & 34.84 & 42.11  \\
    DROP  & 19.09 & 23.16 & 40.23 & 48.03  & - & - & 0.0 & 0.0 & 16.97 & 21.65  \\
    SQuAD  & 68.03 & 78.55 & 56.25 & 64.58  & 46.74 & 57.97 & 0.0 & 0.0 & 56.53 &  71.25 \\
    ROPES  & 40.03 & 49.07 & 24.08 & 31.74  & - & - & - & - & 14.05 & 19.12  \\
    DuoRC  & 25.65 & 34.28 & 27.27 & 34.19 & 30.30 & 35.23 & - & - & 21.51 & 28.85  \\
    \bottomrule
    \end{tabular}
    \end{minipage} \vfill
    \begin{minipage}{\textwidth}
    \centering
    \begin{tabular}{p{2cm}p{5cm}p{5cm}}
    \\
    {\bf Template Type} &   {\bf Answered Incorrectly} & {\bf Answered Correctly} \\
     \midrule
     Invert Choice &  {\color{teal}Original:} Which art gallery was founded first, Art Euphoric or Trescott Street? \hspace{4cm} {\color{olive}Generated:} Which art gallery was founded first, Trescott Street or Art Euphoric?  & {\color{teal}Original:} Who scored more field goals, Nate Kaeding or Dan Carpenter? \hspace{4cm} {\color{olive}Generated:} Who scored more field goals, Dan Carpenter or Nate Kaeding? \\
     \midrule
     Implication  &  {\color{teal}Original:} When did the Huguenots secure the right to own land in the Baronies? \hspace{4cm} {\color{olive}Generated:} Who secured the right to own land in baronies in 1697? & {\color{teal}Original:} When did Henry issue the Edict of Nantes?  \hspace{4cm} {\color{olive}Generated:} What did Henry issue in 1598?\\
     \midrule
     SEARs & {\color{teal}Original:} What was the theme of Super Bowl 50? \hspace{4cm} {\color{olive}Generated:} So what was the theme of Super Bowl 50? & {\color{teal}Original:} Who won Super Bowl 50?  \hspace{4cm} {\color{olive}Generated:} So who won Super Bowl 50? \\
     \bottomrule
    \end{tabular}
    \end{minipage}
\caption{Quantitative and qualitative analysis of generated augmentations. We only show performance for high yield and high-quality augmentations. }
\label{tab:error_analysis}
\end{table*}

\section{Related Work}
\paragraph{Generalization and multi-dataset evaluation}
Recently there has been some work aimed at exploring the relation and differences between multiple reading comprehension datasets.

\textsc{MultiQA} \cite{Talmor2019MultiQAAE} investigates over ten RC  datasets, training on one or more source RC datasets, and evaluating generalization, as well as transfer to a target RC dataset. This work analyzes the factors that contribute to generalization, and shows that training on a source RC dataset and transferring to a target dataset substantially improves  performance. MultiQA also provides a single format including a model and infrastructure for training and comparing question answering datasets. We provide no training mechanism, instead focusing on very simple evaluation that is compatible with any training regime, including evaluating on hidden test sets.

\textsc{MRQA19}, the Machine Reading for Question Answering workshop, introduced a shared task, which tests whether existing machine reading comprehension systems can generalize beyond the datasets on which they were trained. The task provides six large-scale datasets for training, and evaluates generalization to ten different hidden test datasets. However these datasets were modified from there original version, and context was limited to 800 tokens. In addition this shared task only tests for generalization with no intra-domain evaluation.  In contrast, our evaluation server simply provides a single-model evaluation on many different datasets, with no prescriptions about training regimes.

\paragraph{NLP evaluation benchmarks}

The General Language Understanding Evaluation benchmark or \textsc{GLUE} \cite{wang-etal-2018-glue} is a tool for evaluating and analyzing the performance of models across a diverse range of existing NLU tasks. A newer version, Super-GLUE \cite{wang2019superglue} is styled after GLUE with a new set of more difficult language understanding tasks. In this line of work another standard  toolkit for evaluating the quality of universal sentence representations is \textsc{SentEval} \cite{conneau2018senteval}. Similar to \textsc{GLUE}, \textsc{SentEval} also encompasses a variety of tasks, including binary and multi-class classification, natural language inference and sentence similarity.  We differ from \textsc{GLUE} and \textsc{SentEval} by focusing on reading comprehension tasks, and only evaluating a single model on all datasets, instead of allowing the model to be tuned to each dataset separately.

\paragraph{Evaluation Platforms and Competitions in NLP}

The use of online evaluation platform with private test labels has been exercised by various leaderboards on Kaggle and CodaLab, as well as shared tasks at the SemEval and CoNLL conferences.

Additional benchmarks such as \textsc{ParlAI} \cite{miller2017parlai} and \textsc{bAbI} \cite{weston2015towards} proposed a hierarchy of tasks towards building question answering and reasoning models and language understanding.  However these frameworks do not include a standardized evaluation suite for system performance nor do they offer a wide set of reading comprehension tasks.


\section{Conclusion}
We have presented ORB, an open reading benchmark designed to be a comprehensive test of reading comprehension systems, in terms of their generalizability, understanding of various natural language phenomenon, capability to make consistent predictions, and ability to handle out-of-domain questions. This benchmark will grow over time as more interesting and useful reading comprehension datasets are released.  We hope that this benchmark will help drive research on general reading systems.
\section{Acknowledgments}
We would like to thank NFS grant CNS-1730158 for providing computational resources (GPUs) to carry out all the experiments.

\bibliography{bib}
\bibliographystyle{acl_natbib}

\end{document}